\documentclass{opt2026} 

\usepackage{booktabs}

\title[Multiplicative vs.\ Additive State Coupling in Learned
Optimizers]{Multiplicative vs. Additive State Coupling in\\ Learned Optimizer Schedules}

\optauthor{%
\Name{M. Harper Langston} \Email{hlangsto@qti.qualcomm.com}\\
\Name{Pierre-David Letourneau} \Email{pletourn@qti.qualcomm.com}\\
\Name{Dalton Jones} \Email{daltjone@qti.qualcomm.com}\\
\Name{Roland Memisevic} \Email{rmemisev@qti.qualcomm.com}\\
\Name{Peter Scott} \Email{peter.scott@nyu.edu}\\
\Name{Mingu Lee} \Email{mingul@qti.qualcomm.com}\\
\Name{Richard A. Lethin} \Email{rlethin@qti.qualcomm.com}\\
\Name{Christopher Lott} \Email{clott@qti.qualcomm.com}\\
\Name{Fatih Porikli} \Email{fporikli@qti.qualcomm.com}\\
\Name{Harris Teague} \Email{hteague@qti.qualcomm.com}\\
\addr Qualcomm AI Research}

\begin{document}

\maketitle

\begin{abstract}
We learn schedules for a continuous penalty relaxation of MAX-CUT by backpropagating through an unrolled classical solver. A state-blind schedule generalizes to instances substantially larger than those used for training. A state-adaptive bilinear schedule also generalizes out of distribution, outperforming its untrained baseline across 15 seeds ($t=4.1$)\footnote{$t$-statistic~\citep{student1908probable} measures estimated difference relative to standard error; sign indicates the direction of the difference, and larger absolute values indicate stronger evidence against a zero mean difference.  More details in Appendix~\ref{app:statistical_tests}.}. By contrast, an additive ablation with identical inputs and output heads is statistically indistinguishable from the bilinear model in distribution but performs worse than its own baseline out of distribution ($t=-11.7$). This separation appears without regularization and persists across a $30$ round weight-decay sweep: the bilinear advantage remains significant at every tested value, whereas the additive model never recovers. The results in this setting isolate how solver state enters the schedule, rather than state adaptivity alone, as the architectural distinction associated with out-of-distribution transfer.
\end{abstract}

\section{Introduction}
\label{sec:intro}
The MAX-CUT problem can be relaxed continuously through {\emph spins}
$x_i=\tanh(u_i)\in(-1,1)$ and an Ising objective \citep{lucas2014ising}
$\Phi(u;J,\rho)$; this objective combines a soft cut term with a penalty (of strength
$\rho$), which drives $x_i$ toward $\pm1$ during optimization. An instance of a penalty/augmented-Lagrangian method could involve solving
this relaxation by gradient descent (GD) under a continuation schedule that
increases $\rho$ over $K$ stages; such schedules (i.e., how quickly $\rho$ grows
and what step size $\alpha$ each stage uses) are typically hand-tuned. In this work, we 
instead \emph{learn} the schedule, unrolling the $K$-stage solve and backpropagating
through the full trajectory~\citep{andrychowicz2016}.

The simplest schedule learns one $(\rho_k,\alpha_k)$ pair per stage ($2K$ free scalars for $K$ stages), independent of the current iterate; i.e., the schedule is state-\emph{blind}, so it
knows the stage but not whether this instance is already close to binary. An alternative schedule is state-\emph{adaptive}; that is, its update is a function of the current state. We study two ways of incorporating the state into the learned schedule:  an \emph{additive} and a \emph{multiplicative} (bilinear) scheme matching the minimal sufficient structure Ebrahimi and Memisevic \citep{ebrahimi2025} identify for non-trivial state tracking in RNNs.\footnote{Our design constraint is chosen to motivate the architecture and study, not a claim that our solver instantiates~\citep{ebrahimi2025}'s framework or predicts these results.}

\textbf{Contribution.} The state-adaptive bilinear schedule generalizes out of distribution, outperforming its own untrained baseline on instances far beyond the training size range, with statistical significance across $15$ seeds. By contrast, the capacity-matched additive ablation, despite being statistically indistinguishable from the bilinear model in distribution, performs substantially worse than its own baseline out of distribution. These opposing trends are already present without regularization and remain unchanged across a $30$-round sweep of the weight-decay coefficient. Invariant to regularization strength, this architectural distinction is our central empirical finding.

\textbf{Related work.} Our use of ``learning'' differs from both classical MAX-CUT relaxations and recent learned solvers.  Classical methods either round a fixed semidefinite relaxation~\citep{goemans1995maxcut} or directly optimize a nonconvex penalty formulation (as in our setting) \emph{but without learning}. Recent methods use graph neural networks to predict the solver's \emph{output}; e.g., a cut or an SDP-feasible solution, unrolling message passing over the Burer-Monteiro relaxation~\citep{yau2024optgnn}, or predicting a primal-dual certificate end-to-end~\citep{chen2026neuralsdp}. We instead preserve the classical solver and learn its \emph{hyperparameter schedule}.  As a result, the question is not {\emph if} a learned solver produces better cuts than a classical one, but{\emph  whether} the schedule's architectural ``inductive bias'' generalizes beyond its training distribution. More specifically, we consider state-blind and state-adaptive schedules and, moreover,investigate how the solver state is incorporated in the latter case.

\textbf{Connection to frontier optimizers.} Our study focuses on the point where an optimizer conditions its updates on a running summary.  Specifically, we do not learn a MAX-CUT solution directly; instead, we unroll the full sequence of stages in a classical penalty-continuation solver and \emph{learn the hyperparameter schedule that controls it}. Comparing state-blind, bilinear, and additive state-adaptive schedules allows us to vary whether the schedule depends on the solver state and how state enters the schedule (while leaving the underlying solver unchanged). The connection to frontier optimizers comes conceptually with how the choice is made between Adam's coordinate-wise treatment of gradient statistics and the structure-preserving treatments used by Shampoo, K-FAC, and Muon \citep{gupta2018shampoo,martens2015kfac,jordan2024muon}. This analogy is discussed in greater detail in Section~\ref{sec:discussion}, together with a discussion of what our MAX-CUT experiments can establish about these more general optimization methods.  In Section~\ref{sec:method}, we discuss our method and how we train and study the schedule options.

\section{Method} 
\label{sec:method} 
While we provide a brief summary of how we learn the hyperparameter schedule of a classical penalty-continuation solver for MAX-CUT (rather than learning a cut or an update rule directly), more details on the method are provided in Appendix~\ref{app:method_details}.  For a relaxed spin vector $s(u)=\tanh(u)$ and graph adjacency matrix $W$, the solver minimizes 
\begin{equation} 
\Phi(u;W,\rho) = -\widehat{\mathrm{cut}}(u) + \rho\,\mathrm{bviol}(u), 
\label{eq:penalized_objective} 
\end{equation} 
where 
\begin{equation} 
\widehat{\mathrm{cut}}(u) = \frac{1}{2} - \frac{1}{4|E|} \tanh(u)^\top W\tanh(u), \qquad \mathrm{bviol}(u) = \frac{1}{n}\sum_{i=1}^{n} \left(\tanh(u_i)^2-1\right)^2. 
\label{eq:soft_cut_bviol} 
\end{equation} 
The first term rewards high-value cuts, while the second term drives the relaxed spins toward $\{-1,+1\}$. The solver runs for $K$ continuation stages, performing $T$ gradient-descent steps at fixed step size $\alpha_k$ and penalty $\rho_k$ within each stage. Increasing the penalty across stages gradually shifts the solver from improving the relaxed cut toward producing a discrete spin assignment. 

\paragraph{Schedule architectures.} 
We compare three learned schedules. The state-blind baseline assigns a separate step size and penalty increment to each stage, 
\begin{equation} 
\alpha_k=\mathrm{softplus}(a_k), \qquad \rho_{k+1} = \rho_k+\mathrm{softplus}(d_k), \footnote{$\mathrm{softplus}(z)=\log(1+e^z)$.}
\label{eq:state_blind} 
\end{equation} 
using $2K$ learned scalars. It can learn a stagewise continuation rule, but it cannot respond to the current trajectory. 
The two state-adaptive schedules receive the size-independent context 
\begin{equation} 
x^{(k)} = \left[ \rho_k/\rho_0,\; \mathrm{bviol}\!\left(u^{(k)}\right),\; \widehat{\mathrm{cut}}\!\left(u^{(k)}\right),\; k/K \right]^\top \in\mathbb{R}^4. 
\label{eq:context} 
\end{equation} 
The bilinear schedule combines this context with the current penalty through the factored interaction 
\begin{equation} 
f_{\mathrm{bil}}^{(k)} = \left(W_{fx}x^{(k)}\right) \odot \left(w_{f\rho}\log\rho_k\right), 
\label{eq:factor} 
\end{equation} 
following the multiplicative construction of \citet{sutskever2011}. 
The normalized ratio $\rho_k/\rho_0$ describes relative continuation progress, while $\log\rho_k$ provides a compressed representation of the absolute penalty scale to the multiplicative gate. Two softplus heads map $f_{\mathrm{bil}}^{(k)}$ to $\alpha_k$ and the nonnegative increment $\rho_{k+1}-\rho_k$, namely,
\begin{equation} 
\alpha_k = \mathrm{softplus} \left( w_\alpha^\top f_{\mathrm{bil}}^{(k)} + b_\alpha \right), \qquad \rho_{k+1} = \rho_k + \mathrm{softplus} \left( w_\rho^\top f_{\mathrm{bil}}^{(k)} + b_\rho \right). 
\label{eq:app_bilinear_heads_body} 
\end{equation} 
To determine whether performance depends on state adaptation alone or on the form of the state interaction, we compare against a capacity-matched additive ablation, 
\begin{equation} 
f_{\mathrm{add}}^{(k)} = \tanh\!\left( W_{\mathrm{add}} [x^{(k)};\log\rho_k] + b_{\mathrm{add}} \right). 
\label{eq:additive} 
\end{equation} 

\begin{figure}[thbp]
\centering
\includegraphics[width=0.75\columnwidth]{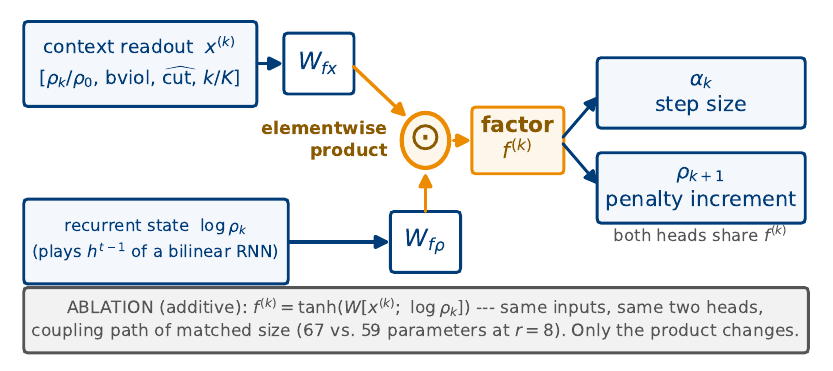}
\caption{\small State-adaptive schedule architectures. The bilinear schedule projects the context and current penalty separately, then combines them by element-wise multiplication. The additive ablation concatenates the same inputs and applies an affine map followed by $\tanh$. Both architectures use the same output heads to produce the step size $\alpha_k$ and penalty increment $\rho_{k+1}-\rho_k$.}
\label{fig:gate}
\end{figure}

The additive model receives the same inputs and uses the same output heads as in Eq.~\eqref{eq:app_bilinear_heads_body} but replaces multiplicative conditioning with a concatenate-and-project layer. Comparing either state-adaptive model with the state-blind baseline tests {\emph whether observing the solver state is useful}; comparing the bilinear and additive models tests {\emph whether the form of that conditioning matters}. Figure~\ref{fig:gate} summarizes these two state-adaptive architectures.

\paragraph{Training.}
We unroll the complete $K$-stage solve and backpropagate through every
inner update and stage transition, following the learning-to-learn
approach of \citet{andrychowicz2016}. The inner loop performs gradient
descent on the MAX-CUT relaxation to update the spins $u$, while an
outer Adam optimizer updates the schedule parameters using the
batch-averaged objective:
\[
\mathcal{L}
=
-\mathbb{E}\!\left[
\widehat{\mathrm{cut}}(u_{\mathrm{final}})
\right].
\]
Because the entire $K$-stage solve is differentiable, gradients
propagate through all $KT$ inner updates and schedule decisions.
Parameters $W_{fx}$ and $w_{f\rho}$ are learned through
their effect on generated schedules $(\alpha_k,\rho_k)$ and the quality of the final solution.\footnote{Since the schedule outputs $(\alpha_k,\rho_k)$ directly affect every inner gradient-descent update, the final loss is differentiable with respect to $W_{fx}$, $w_{f\rho}$, and the remaining schedule parameters, optimized by Adam through backpropagation.} Spin
variables $u$ are instance-specific
optimization states (not learned) produced by the inner solver. All architectures
share the same objective and gradient-descent solver and differ only in schedule parameterization.  Details in Section~\ref{app:method_details}.

\section{Experiments}
\label{sec:experiments}

All schedules are trained for 200 epochs on freshly sampled
Erd\H{o}s--R\'enyi $G(n,0.5)$ graphs~\cite{erdos1959random} with
$n\in\{6,\ldots,15\}$, using $K=7$ continuation stages, $15$ inner
gradient steps per stage, and batch size of $8$. Each architecture is
compared with its own frozen, untrained counterpart, which uses the same
architecture but retains its initial geometric penalty ladder and
constant step size.

We evaluate out of distribution on synthetic $G(n,0.5)$ graphs with
$n\in\{30,50,100\}$ and on the 80-instance Biq Mac suite with $n=100$
and known optima \citep{wiegele2007biqmac}. As the synthetic
instances have no known optimum, we report discrete cut values. On
Biq Mac, we report the ratio of the discrete cut to the known optimum.
All results aggregate 15 independent
train-and-evaluate runs.

\paragraph{In-distribution performance.} The three trained schedules perform similarly on held-out training-size graphs: their mean discrete cuts range from $19.35$ to $19.47$, compared with $18.72$ for the state-blind schedule's frozen baseline. 
On a fixed hard instance with $n=15$ and optimum $38$, the additive, bilinear, and state-blind schedules reach the optimum in $13/15$, $12/15$, and $7/15$ runs, respectively. 
Thus, the bilinear and additive schedules cannot be reliably distinguished in distribution. Appendix~\ref{app:indist} reports the complete held-out results, frozen-baseline comparisons, and seed-matched analysis of the fixed hard instance. A visual comparison of the in-distribution and out-of-distribution evaluations appears in Figure~\ref{fig:bar_main}.

\paragraph{Out-of-distribution performance.}

Table~\ref{tab:generalization} evaluates schedules trained only on
$n\leq15$ on instances with $n=100$. On Biq Mac, the state-blind
schedule improves over its own frozen baseline by $0.0161$, while the
bilinear schedule improves by $0.0084$ (paired $t=4.1$ over 15 seeds
\citep{student1908probable}). The additive schedule instead declines by
$0.0522$ relative to its baseline ($t=-11.7$), despite being
statistically indistinguishable from the bilinear schedule in
distribution.

These changes are relative to architecture-specific baselines and not to be confused with absolute performance. The state-blind model begins
from a substantially weaker baseline than the bilinear model
($0.9259$ versus $0.9543$), so its larger relative improvement still
produces a lower final score ($0.9420$ versus $0.9627$). The bilinear
schedule is therefore the strongest trained model in the table.

\begin{table}[t]
\small
\centering
\caption{\small
  Out-of-distribution performance without regularization over
  \textbf{15 seeds}. Synthetic $n=100$ results are discrete cut values
  (mean$\pm$standard deviation); Biq Mac results are ratios to known
  optima over 80-instance suite \citep{wiegele2007biqmac}. ``Frozen''
  denotes each architecture's untrained schedule, and ``learned'' denotes
  its trained schedule. The final row reports a common geometric schedule
  selected by grid search on the training distribution.
}
\label{tab:generalization}
\begin{tabular}{l@{\hskip 10pt}cc@{\hskip 12pt}cc}
\toprule
& \multicolumn{2}{c@{\hskip 12pt}}{$n=100$ synthetic (discrete cut)}
& \multicolumn{2}{c}{Biq Mac ratio to optimum} \\
\cmidrule(r{10pt}){2-3}
\cmidrule{4-5}
Model & frozen & learned & frozen & learned \\
\midrule
State-blind (scalar)
  & $1393.1\pm6.7$ & $1406.9\pm7.4$
  & $0.9259\pm0.0030$ & $0.9420\pm0.0053$ \\
Bilinear
  & $1410.9\pm6.8$ & $1415.4\pm7.3$
  & $0.9543\pm0.0024$ & $0.9627\pm0.0078$ \\
Additive (ablation)
  & $1411.0\pm6.8$ & $1397.9\pm12.3$
  & $0.9545\pm0.0024$ & $0.9023\pm0.0166$ \\
\midrule
\multicolumn{3}{l}{\emph{Tuned fixed schedule} (common reference)}
  & \multicolumn{2}{c}{$0.9624\pm0.0024$} \\
\bottomrule
\end{tabular}
\end{table}

\paragraph{Comparison with a tuned fixed schedule.} The frozen schedules are intentionally untuned, so we also compare against a stronger fixed reference.
We grid-search a geometric schedule of the form $\rho_k=\rho_0g^k$ with constant $\alpha$, selecting its parameters exclusively on the training distribution. The selected schedule obtains a Biq Mac ratio of $0.9624$, exceeding the learned state-blind and additive schedules. The unregularized bilinear result, $0.9627$, is statistically indistinguishable from this tuned reference. 
Appendix~\ref{app:wdsweep} gives the complete 150-configuration search protocol, the selected hyperparameters, and the ranking of the untuned frozen initialization.

\begin{figure}[t]
\centering
\includegraphics[width=0.45\textwidth]{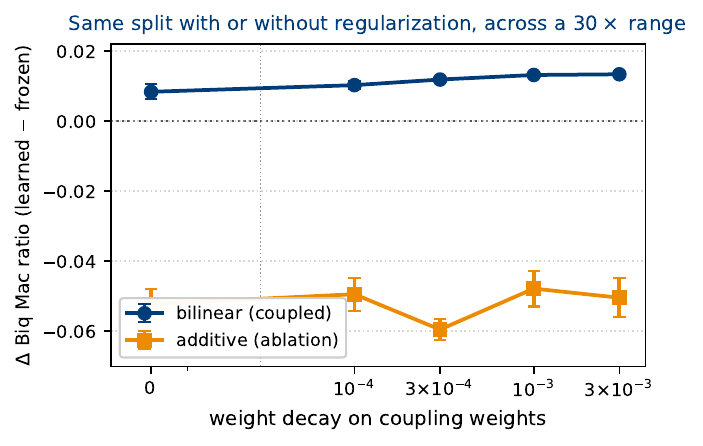}
\caption{\small
  Change in Biq Mac ratio after training, relative to each architecture's
  frozen baseline, across the weight-decay sweep. Bilinear improvement
  remains positive, whereas additive change remains negative,
  including at zero weight decay. Complete numerical
  results reported in Appendix~\ref{app:wdsweep}.
}
\label{fig:sweep}
\end{figure}

\paragraph{Robustness to weight decay.} The bilinear-additive separation is already present without regularization and persists across the full weight-decay sweep. 
As seen in Figure~\ref{fig:sweep}, with $\mathrm{w_{dec}}\in \{0,10^{-4},3\times10^{-4},10^{-3},3\times10^{-3}\}$ the bilinear improvement over its frozen baseline remains positive, ranging from $+0.0084$ to $+0.0134$; meanwhile, the additive change remains negative, ranging from $-0.0595$ to $-0.0478$ at nonzero decay and equaling $-0.0522$ at zero decay.
Appendix~\ref{app:wdsweep} reports the full per-value results, paired tests, seed-wise signs, between-architecture comparisons, and fixed-instance optimum rates.

\paragraph{Summary.} The bilinear and additive schedules have similar in-distribution performance but diverge beyond the training size range. 
The bilinear model improves over its frozen baseline and matches the tuned fixed reference without regularization; however, the additive model degrades substantially. 
Appendix~\ref{app:indist} shows that this separation is not visible on held-out training-size graphs or the single fixed hard instance. Appendix~\ref{app:wdsweep} further shows that this is not explained by an untuned reference schedule or by the choice of weight-decay strength.

\section{Discussion and Limitations}
\label{sec:discussion}

Our bilinear schedule was motivated by \citet{ebrahimi2025}'s state-tracking
observations, treated here as architectural motivation rather than a
mechanistic explanation. Nonetheless, a concrete hypothesis is that
$f_{\mathrm{bil}}^{(k)}$ is an \emph{unsaturated} elementwise product in
which $\rho_k/\rho_0$ and $\log\rho_k$ pass through unclamped and compound
across stages, while the additive path's $\tanh$ self-limits the same
inputs. This is exactly the asymmetry that lets bilinear (not additive)
coupling grow its penalty schedule without bound in a related setting. While we have not re-verified this
trajectory-level effect for the MAX-CUT setting studied here, we hypothesize that our comparison varies activation geometry
together with multiplicative structure.  Disentangling the two (e.g., via a saturating bilinear head or an unsaturated linear additive head) is a natural next ablation for future work.

\paragraph{Connection to learned optimizers.}
The architectural choice studied here, \emph{how} state enters an update,
not merely whether it does, arises whenever an optimizer conditions on a
running summary. We do not claim the mechanism observed here transfers to
methods such as Adam, Shampoo, K-FAC, or Muon~\citep{gupta2018shampoo,martens2015kfac,jordan2024muon}, but it motivates an
analogous controlled comparison in those families.



\paragraph{Scope and limitations.}
The observed separation persists against a tuned fixed schedule, so our
contribution is architectural: differential transfer between two
matched-capacity state-adaptive forms; specifically, this is not a claim that learning
surpasses careful manual tuning in this small-instance regime. The evidence
is specific to one MAX-CUT relaxation, one continuation solver, one
training distribution, and one evaluation suite;
Appendix~\ref{app:untested_axes} lists the untested generalization axes and
reports a knapsack extension that shares the same structural asymmetry
(unsaturated bilinear coupling vs.\ saturating additive coupling) without a
matching out-of-distribution quality separation, evidence that the
asymmetry itself, though not its transfer consequence, {\emph can} extend beyond
MAX-CUT. Appendix~\ref{app:statistical_tests} discusses a further
limitation of the statistical evidence itself.



\section{Conclusion}

We investigated the learning of continuation schedules for a classical penalty relaxation of
MAX-CUT and found that out-of-distribution behavior depends on how the
schedule incorporates solver state. Among two state-adaptive schedules
with identical inputs, output heads, and training procedure, the bilinear
model improves significantly over its frozen baseline while the
capacity-matched additive ablation degrades substantially, isolating the
\emph{form} of state coupling and not merely state-adaptivity, as the
architectural variable behind this separation. A preliminary knapsack
extension shares the same coupling asymmetry but not the same quality
separation, leaving open whether the effect depends on the constraint's
dual structure, the solver, or the mechanism itself. We are currently testing our activation-geometry hypothesis more directly with a saturating bilinear
head and an unsaturated additive head abalation study, and we plan to report on our findings in the near future.

\bibliography{biblio}

\newpage
\appendix

\section{Detailed Method and Training Setup} \label{app:method_details} 
This appendix gives the complete solver, schedule, and training specification underlying Section~\ref{sec:method}. 
\subsection{MAX-CUT relaxation} 
Let $G=(V,E)$ be an undirected graph with adjacency matrix $W\in\mathbb{R}^{n\times n}$. 
A discrete spin vector $s\in\{-1,+1\}^n$ induces the normalized cut value 

\begin{equation} 
\mathrm{cut}(s) = \frac{1}{2} - \frac{1}{4|E|}s^\top W s. 
\label{eq:app_discrete_cut} 
\end{equation} 

We parameterize a relaxed spin vector as 
\begin{equation} 
s(u)=\tanh(u), \qquad u\in\mathbb{R}^n,
\end{equation} 
defining the corresponding normalized soft-cut value by 
\begin{equation} 
\widehat{\mathrm{cut}}(u) = \frac{1}{2} - \frac{1}{4|E|} \tanh(u)^\top W\tanh(u). 
\label{eq:app_soft_cut} 
\end{equation} 

The binary-violation statistic 
\begin{equation} 
\mathrm{bviol}(u) = \frac{1}{n} \sum_{i=1}^{n} \left(\tanh(u_i)^2-1\right)^2 
\label{eq:app_bviol} 
\end{equation} 
measures the distance of the relaxed spins from the saturated values $\{-1,+1\}$. 
At penalty level $\rho$, the solver minimizes 

\begin{equation} \Phi(u;W,\rho) = -\widehat{\mathrm{cut}}(u) + \rho\,\mathrm{bviol}(u). 
\label{eq:app_penalized_objective} 
\end{equation} 
While the negative soft-cut term favors high-value cuts, the penalty term encourages the relaxed solution to become discrete.

\begin{figure}[h]
\centering
\includegraphics[width=0.95\textwidth]{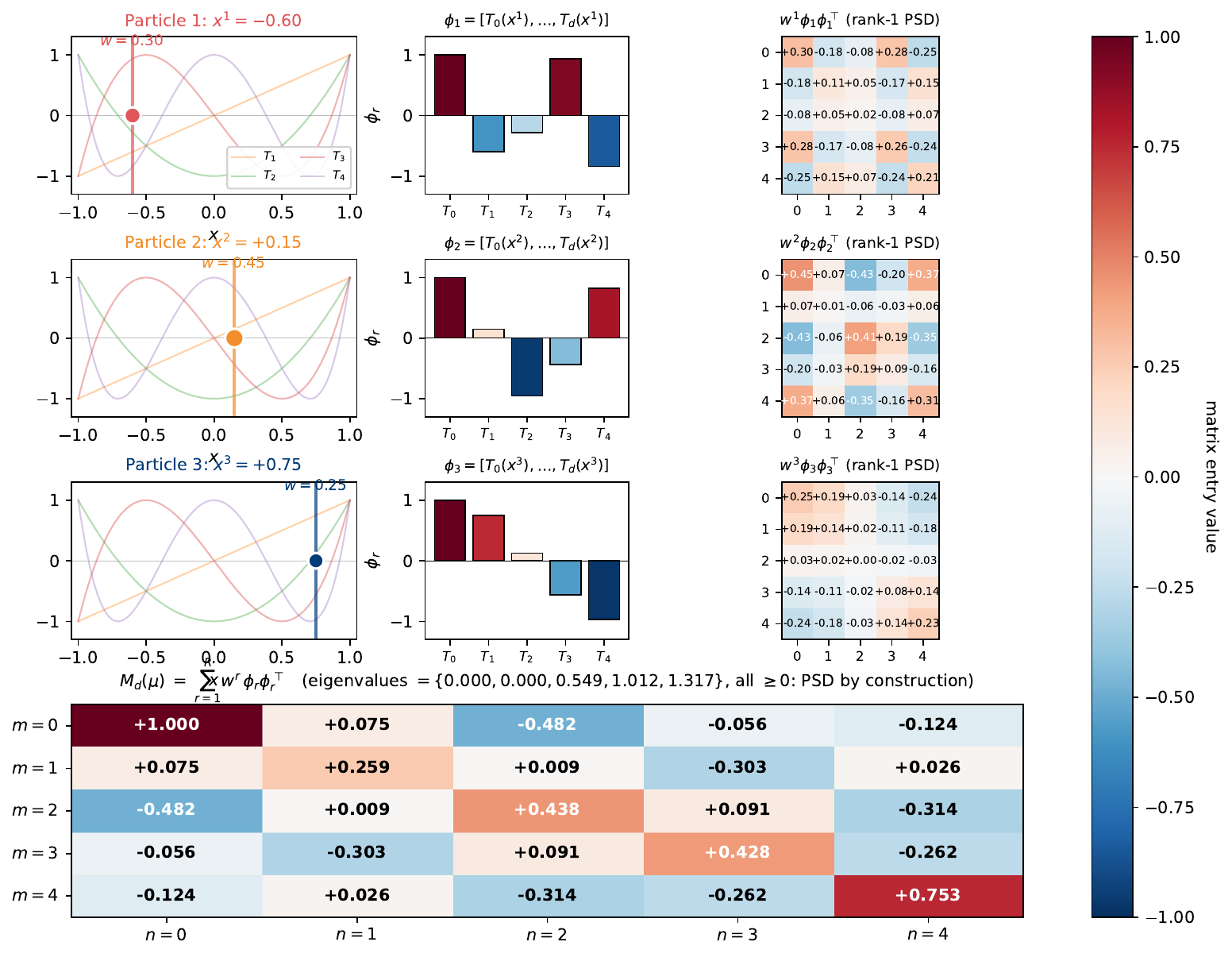}
\caption{PSD-by-construction of the moment matrix, shown for a general $R{=}3$-atom marginal (this paper's MAX-CUT relaxation is the $R{=}1$ special case). \textbf{Rows:} three atoms at locations $x^{r}$ with weights $w^{r}$; their Chebyshev basis vectors $\phi_r=[T_0(x^{r}),\dots,T_d(x^{r})]$; and the weighted rank-one outer products $w^{r}\phi_r\phi_r^\top$. \textbf{Bottom:} their sum $M_d(\mu)=\sum_r w^{r}\phi_r\phi_r^\top$ (Eq.~\ref{eq:moment}), whose eigenvalues are all nonnegative. A sum of nonnegative-weighted rank-one terms is PSD for any atoms and weights, so no semidefinite constraint is enforced during optimization.}
\label{fig:psd}
\end{figure}

\begin{figure}[h]
\centering
\includegraphics[width=0.72\textwidth]{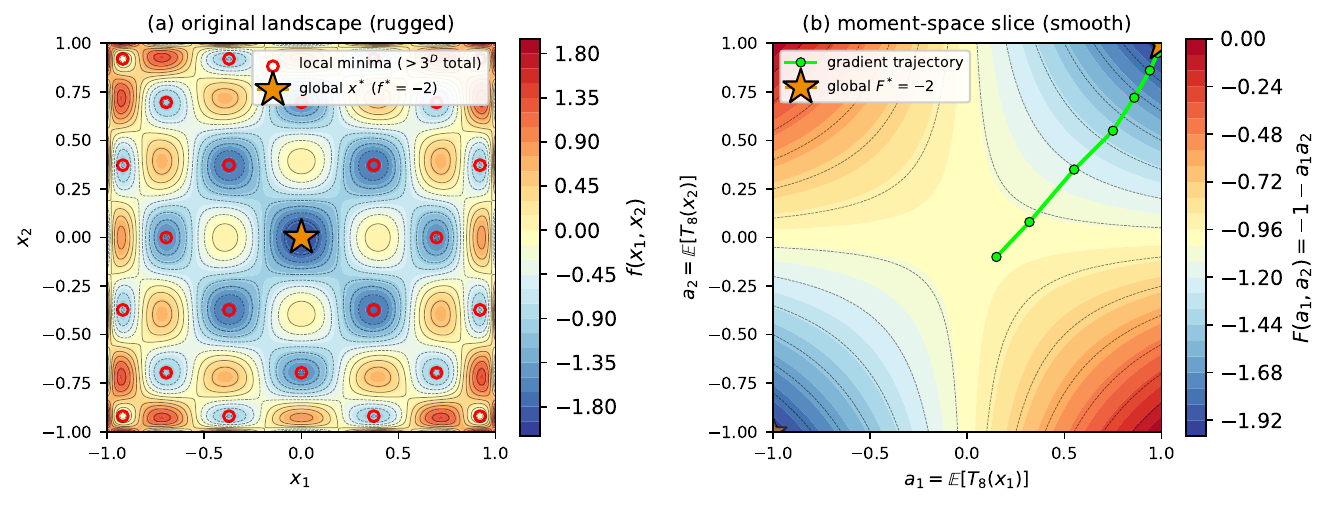}
\caption{The particle-parameterization solver trades a rugged original landscape (left; red circles mark a sample of its $>3^D$ local minima) for a smooth moment-space landscape (right) with a single global minimum reachable by gradient-based search (green trajectory), shown on an illustrative 2D objective unrelated to MAX-CUT. This paper's MAX-CUT relaxation uses the rank-one ($R{=}1$) case of the same parameterization.}
\label{fig:flatten}
\end{figure}

\paragraph{Relation to the general particle relaxation.}
The relaxation above is the single-atom special case of the particle-parameterized moment relaxation for general polynomial optimization~\citep{letourneau2023polyopt}. That framework rewrites a polynomial program as an infimum of $\mathbb{E}_\mu[p(x)]$ over probability measures $\mu$ on $[-1,1]^D$ subject to a positive-semidefinite moment-matrix condition, and represents $\mu$ by a mixture of product measures. Each coordinate carries $R$ atoms at locations $x^{r}=\tanh(u^{r})$ with softmax weights $w^{r}\ge0$, $\sum_r w^{r}=1$. Lifting each atom into the degree-$d$ Chebyshev basis $\phi(x)=[T_0(x),\dots,T_d(x)]^\top$, the marginal moment matrix of one coordinate is
\begin{equation}
  M_d(\mu) \;=\; \sum_{r=1}^{R} w^{r}\,\phi(x^{r})\,\phi(x^{r})^\top ,
  \label{eq:moment}
\end{equation}
a nonnegative combination of rank-one outer products. It is therefore PSD for \emph{any} atom locations and \emph{any} nonnegative weights, so no semidefinite constraint or projection is ever imposed during optimization (Figure~\ref{fig:psd}). The motivation is geometric: the original objective is rugged, with a number of local minima that grows exponentially in the dimension, whereas the moment-space objective is smooth enough for gradient-based search to reach a global minimum (Figure~\ref{fig:flatten}).

MAX-CUT on $n$ nodes is the rank-one case $R{=}1$: one atom per node with weight $w\equiv1$, so a coordinate reduces to the single spin $s_i=\tanh(u_i)\in(-1,1)$ and Eq.~\eqref{eq:moment} collapses to $\phi(s_i)\phi(s_i)^\top$, trivially PSD. The moment machinery is thus present but never binding, there is no weight vector to train, and the only optimization variables are the spins $u\in\mathbb{R}^n$. This is why the objective~\eqref{eq:app_penalized_objective} contains no moment-matrix term and why the learned schedule acts entirely on the per-stage step size $\alpha_k$ and penalty $\rho_k$ of this spin-space problem. We state the general construction only to place the relaxation; none of the claims in this paper rest on $R>1$.

\subsection{Continuation solver}
The solver uses $K=7$ continuation stages. 
At stage $k$, the penalty $\rho_k$ and step size $\alpha_k$ remain fixed for $T=15$ gradient-descent updates: 
\begin{equation} 
u_{t+1}^{(k)} = u_t^{(k)} - \alpha_k \nabla_u \Phi\!\left(u_t^{(k)};W,\rho_k\right), \qquad t=0,\ldots,T-1. 
\label{eq:app_inner_update} 
\end{equation} 

The final iterate of each stage initializes the next, 
\begin{equation} 
u_0^{(k+1)} = u_T^{(k)}. 
\label{eq:app_stage_transition} 
\end{equation} 

The solver follows one continuous trajectory across stages rather than restarting after each change in the penalty. Continuation separates the relaxed search from the eventual enforcement of discrete spins. At lower penalties, the solver can improve the soft-cut value without being forced immediately toward a particular binary assignment. Increasing the penalty gradually makes fractional spins more costly. If the penalty rises too quickly, the trajectory can commit to a poor discrete configuration before finding a good relaxed cut. If it rises too slowly, the final iterate can remain insufficiently discrete. The step size controls progress within each stage and must likewise balance slow convergence against unstable updates. 

\subsection{Context representation} 
The full iterate $u^{(k)}$ has dimension $n$ and therefore cannot be passed directly to a schedule whose architecture is independent of graph size. We instead summarize the current solve using 
\begin{equation} 
x^{(k)} = 
\left[ \frac{\rho_k}{\rho_0},\; 
\mathrm{bviol}\!\left(u^{(k)}\right),\; 
\widehat{\mathrm{cut}}\!\left(u^{(k)}\right),\; 
\frac{k}{K} \right]^\top \in\mathbb{R}^4. 
\label{eq:app_context} 
\end{equation} 
The soft-cut and binary-violation terms describe the quality and discreteness of the current iterate. The ratio $\rho_k/\rho_0$ records the relative progress of the penalty continuation, while $k/K$ records the fraction of stages completed. The dimension of this representation does not depend on the number of vertices. 

The penalty also enters the state-adaptive architectures as $\log\rho_k$. The two representations have different roles. 
The normalized value $\rho_k/\rho_0$ places continuation progress on a relative scale and appears alongside the other solver-state features. The logarithm represents the current absolute penalty scale while compressing values that may span several orders of magnitude. In the bilinear model, this logarithmic branch modulates the projected context features directly. 

\subsection{Schedule architectures} 
\subsubsection{State-blind schedule} 
The state-blind schedule learns an independent step size and penalty increment for each continuation stage: 
\begin{equation} 
\alpha_k = \mathrm{softplus}(a_k), \qquad \rho_{k+1} = \rho_k+\mathrm{softplus}(d_k), \qquad k=0,\ldots,K-1, 
\label{eq:app_state_blind} 
\end{equation} 
where $\{a_k,d_k\}_{k=0}^{K-1}$ are $2K$ free scalar parameters and 
\begin{equation} 
\mathrm{softplus}(z)=\log(1+e^z). 
\end{equation} 
The schedule can vary with the continuation stage but cannot respond to the graph or the current optimization trajectory. 

\subsubsection{Factored bilinear schedule} 
The bilinear schedule forms a factor vector 
\begin{equation} 
f_{\mathrm{bil}}^{(k)} = \left(W_{fx}x^{(k)}\right) \odot \left(w_{f\rho}\log\rho_k\right) \in\mathbb{R}^r, 
\label{eq:app_bilinear_factor} 
\end{equation} 
where $\odot$ denotes elementwise multiplication, $W_{fx}\in\mathbb{R}^{r\times m}$, $m=4$, and $w_{f\rho}\in\mathbb{R}^r$. 
We use factor width $r=8$ in all experiments. The factorization is linear in each branch before their elementwise product and has a parameter count independent of graph size. 

Two output heads map the factor vector to the step size and penalty increment: 
\begin{equation} \alpha_k = \mathrm{softplus} \left( w_\alpha^\top f_{\mathrm{bil}}^{(k)} + b_\alpha \right), \qquad \rho_{k+1} = \rho_k + \mathrm{softplus} \left( w_\rho^\top f_{\mathrm{bil}}^{(k)} + b_\rho \right). 
\label{eq:app_bilinear_heads} 
\end{equation} 

The softplus heads keep the step size positive and the penalty nondecreasing. The model has $O\!\left(rm\right)$ parameters and contains $59$ parameters at $r=8$ under the parameterization used in our experiments. 

\subsubsection{Additive ablation} 
The additive schedule receives the same inputs but replaces the multiplicative factorization with 
\begin{equation} 
f_{\mathrm{add}}^{(k)} = 
\tanh\!\left( W_{\mathrm{add}} \left[x^{(k)};
\log\rho_k\right] + b_{\mathrm{add}} \right). 
\label{eq:app_additive_factor} 
\end{equation} 

It then substitutes $f_{\mathrm{add}}^{(k)}$ for $f_{\mathrm{bil}}^{(k)}$ in Eq.~\eqref{eq:app_bilinear_heads}. 
The additive model therefore uses the same context, recurrent penalty state, factor width, and output heads as the bilinear model. Only the interaction between the context and penalty changes.

At $r=8$, the additive model contains $67$ parameters, compared with $59$ for the bilinear model. The difference consists of the $r$ hidden biases in the additive layer, which have no counterpart in the bilinear factorization. Thus, the additive model has slightly more, rather than fewer, parameters. 

The three architectures support two controlled comparisons. Comparing either state-adaptive model with the state-blind schedule tests whether conditioning on the current solver state is useful. Comparing the bilinear and additive schedules tests whether the form of that conditioning matters when the available state, output parameterization, training procedure, and approximate capacity are otherwise held fixed. 


\subsection{Unrolled training} We train the schedule parameters by differentiating through the complete $K$-stage solve, following the learning-to-learn formulation of \citet{andrychowicz2016}. The inner gradient-descent updates retain the higher-order computation graph. Gradients from the training objective therefore propagate through all $T$ updates per stage, all stage transitions, and all schedule decisions. 

The continuation stages are part of the training graph, not merely an evaluation procedure. A schedule decision at an early stage changes the iterate passed to the next stage, which in turn changes the next context vector and all later schedule decisions. 
Backpropagation through the unrolled solve evaluates each decision through its cumulative effect on the final trajectory. 

Each training batch contains eight freshly sampled Erd\H{o}s--R\'enyi graphs $G(n,0.5)$~\cite{erdos1959random}, with \begin{equation} n\sim\operatorname{Uniform}\{6,\ldots,15\}. \end{equation} Sampling new graphs in each batch prevents training from relying on a fixed collection of instances. 

For each graph, we evolve $L=8$ independent rank-one spin trajectories under a shared schedule. 
The training objective is averaged over the $L$ replicas during the solve. 
At evaluation, all replicas are discretized and the best cut is reported. 
The shared schedule therefore receives a training signal from several independent trajectories, while evaluation retains the benefit of parallel restarts. 

\subsection{Nested optimization and gradient flow}

Training consists of two nested optimization problems over different
variables. The \emph{inner} optimization solves for the relaxed spins
$u$, while the \emph{outer} optimization learns the schedule parameters.

\paragraph{Inner optimization.}
For a graph instance and a fixed schedule, the solver performs plain
gradient descent on the penalized MAX-CUT objective
$\Phi(u;W,\rho_k)$. At continuation stage $k$, with step size
$\alpha_k$ and penalty $\rho_k$ held fixed, the updates are

\begin{equation}
g_t^{(k)}
=
\nabla_u \Phi\!\left(u_t^{(k)};W,\rho_k\right),
\qquad
u_{t+1}^{(k)}
=
u_t^{(k)}
-
\alpha_k g_t^{(k)}.
\end{equation}

The solver uses $K=7$ continuation stages and $T=15$ gradient updates
per stage. No parameters are learned in this inner loop; it is simply
gradient descent applied to the relaxation objective.

\paragraph{Outer optimization.}
The learned quantities are the schedule parameters,
\[
\theta
=
\{
W_{fx},
w_{f\rho},
w_\alpha,
w_\rho,
b_\alpha,
b_\rho,
\rho_0
\},
\]
which determine the sequence of penalties and step sizes used by the
inner solver. Training minimizes

\begin{equation}
\mathcal L
=
-\widehat{\mathrm{cut}}
\!\left(u_{\mathrm{final}}\right),
\end{equation}

where $u_{\mathrm{final}}$ is the final iterate produced by the complete
$K$-stage solve. Minimizing $\mathcal L$ therefore corresponds to
maximizing the final cut value achieved by the schedule. For each
training iteration, we sample a batch of random graphs, execute the full
unrolled solve, average the losses across the batch, and update
$\theta$ with Adam.

\paragraph{Differentiating through the solver.}
The inner-loop gradients are computed with the higher-order computation
graph retained, so the gradient-descent updates themselves remain
differentiable. Consequently, the final loss depends on the schedule
parameters through the entire unrolled optimization trajectory. For the
bilinear schedule,

\begin{equation}
f_{\mathrm{bil}}^{(k)}
=
\left(W_{fx}x^{(k)}\right)
\odot
\left(w_{f\rho}\log\rho_k\right),
\end{equation}

which determines the schedule outputs $(\alpha_k,\rho_{k+1})$ used by
the inner updates

\begin{equation}
u_{t+1}^{(k)}
=
u_t^{(k)}
-
\alpha_k
\nabla_u
\Phi\!\left(u_t^{(k)};W,\rho_k\right).
\end{equation}

The resulting dependency chain is

\begin{equation}
W_{fx},\,w_{f\rho},\,\ldots
\rightarrow
\{\alpha_k,\rho_k\}_{k=1}^{K}
\rightarrow
u_{\mathrm{final}}
\rightarrow
\mathcal L ,
\end{equation}

so that

\begin{equation}
\frac{\partial \mathcal L}{\partial W_{fx}}
\neq 0,
\qquad
\frac{\partial \mathcal L}{\partial w_{f\rho}}
\neq 0.
\end{equation}

Backpropagation traverses the complete unrolled solve, containing
$KT=105$ gradient-descent updates, and produces gradients with respect
to all schedule parameters. The outer Adam optimizer then updates
$W_{fx}$, $w_{f\rho}$, $w_{\alpha}$, $w_{\rho}$, $b_{\alpha}$,
$b_{\rho}$, and $\rho_0$. Training can therefore be viewed as a
learning-to-learn procedure \citep{andrychowicz2016}, in which the
continuation solver acts as a differentiable computation graph and the
outer optimizer learns the schedule that produces the best final cuts on
average over the training distribution.

\subsection{Training hyperparameters and stability}
The outer-loop optimizer is Adam with learning rate $0.05$. Because the predicted step sizes and penalties affect all later stages, one poor outer-loop update can destabilize the entire unrolled trajectory. We use two safeguards: 
\begin{equation} 
\alpha_k\in[10^{-4},10], \qquad \left\|\nabla_\theta\mathcal{L}\right\|_2\leq 1.0, 
\label{eq:app_stability} 
\end{equation} where $\theta$ denotes the learned schedule parameters. The step-size clamp prevents extremely large inner updates and steps that collapse toward zero. Gradient-norm clipping limits abrupt changes to the schedule parameters during backpropagation through the unrolled solver. 

The principal solver and training hyperparameters are 
\begin{equation} 
K=7, \qquad T=15, \qquad L=8, \qquad r=8, \qquad \eta_{\mathrm{Adam}}=0.05. 
\label{eq:app_hyperparameters} 
\end{equation} 
Here, $K$ and $T$ determine the length of the differentiable solve, $L$ determines the number of parallel trajectories, and $r$ determines the width of the state-adaptive factor representation. These settings are held fixed across the architecture comparison. 

All training and evaluation runs use double precision on the CPU. Each reported seed corresponds to a complete independent training and evaluation run. 

\subsection{Regularization sweep} 
To determine whether the architectural comparison depends on regularization strength, we apply decoupled weight decay~\citep{loshchilov2019decoupled} to all learned parameters and sweep 
\begin{equation} 
\mathrm{w_{dec}} \in \left\{ 0,\; 10^{-4},\; 3\times10^{-4},\; 10^{-3},\; 3\times10^{-3} \right\}. 
\label{eq:app_weight_decay} 
\end{equation} 
We use 15 independent seeds for every architecture and weight-decay setting. The sweep is used to test the robustness of the architectural comparison, rather than to select a separate architecture for each evaluation regime. 

\subsection{How the schedule affects MAX-CUT performance} 
The schedules can affect MAX-CUT performance only through their control of the fixed solver. 
They do not alter the objective, predict an update direction, or produce the final cut directly. 
Instead, they determine how far the solver moves within each stage and how quickly the penalty shifts the trajectory from relaxed cut optimization toward a discrete spin assignment. 

The state-blind schedule learns an average stagewise progression across the training distribution, but it cannot react to the current trajectory. 
The additive schedule can respond to the current soft-cut value, binary violation, stage, and penalty, combining these signals through a standard affine layer. 
The bilinear schedule observes the same state but allows the current penalty scale to modulate the projected context features multiplicatively. 

In distribution, the bilinear and capacity-matched additive schedules are statistically indistinguishable. 
Outside the training size range, the bilinear schedule significantly outperforms its own untrained baseline across 15 independent seeds, including on instances far larger than those used during training. 
The additive schedule instead performs substantially worse than its own untrained baseline out of distribution. 
These opposing trends are already present without weight decay and remain unchanged across the regularization sweep. 
The comparison therefore indicates that access to state alone is insufficient to explain transfer; how the state enters the schedule is also consequential.

\section{Additional In-Distribution Results} \label{app:indist} This section expands the in-distribution evaluation summarized in Section~\ref{sec:experiments}. We report the architecture-specific frozen baselines, the held-out results over the training size range, and the seed-matched outcomes on the fixed hard instance. These experiments provide in-distribution controls for the out-of-distribution comparison in Table~\ref{tab:generalization}.

\paragraph{Held-out random instances.} We first evaluate on held-out Erd\H{o}s--R\'enyi $G(n,0.5)$ graphs with $n\in\{6,\ldots,15\}$, matching the training distribution. 
Across 15 independent training seeds, the mean discrete cut is $19.35$ for the state-blind schedule and lies between $19.44$ and $19.47$ for the two state-adaptive schedules. 
The corresponding frozen baselines obtain $18.72$ for the state-blind architecture and between $19.33$ and $19.34$ for the state-adaptive architectures. 
Thus, all three learned schedules improve over their respective frozen baselines in distribution, while the differences among the trained models remain within the observed run-to-run variation. 
This evaluation primarily verifies that each architecture can learn a useful schedule on the distribution used for training. 

\begin{figure}[thbp]
\centering
\includegraphics[width=0.95\textwidth]{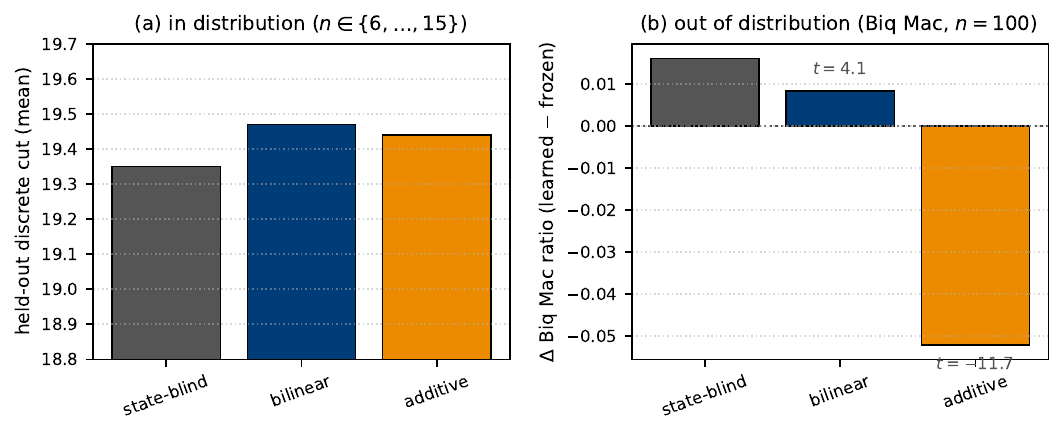}
\caption{\small
  Comparison of in-distribution and out-of-distribution behavior. Left: held-out training-distribution graphs ($n\in\{6,\ldots,15\}$). All three learned schedules perform similarly, with the bilinear and additive models falling within the observed variability. Right: out-of-distribution Biq Mac evaluation ($n=100$), shown as the change relative to each architecture's own frozen baseline ($\Delta = R_{\mathrm{learned}} - R_{\mathrm{frozen}}$). The state-blind and bilinear schedules improve after training ($\Delta=+0.0161$ and $+0.0084$, respectively), whereas the additive ablation degrades substantially ($\Delta=-0.0522$, paired $t=-11.7$). The separation between bilinear and additive schedules is therefore not visible on the training distribution and emerges only under distribution shift.}
\label{fig:bar_main}
\end{figure}

Figure~\ref{fig:bar_main} summarizes the distinction between the in-distribution and out-of-distribution evaluations. On held-out graphs drawn from the training distribution, the three learned schedules achieve nearly identical discrete-cut performance, with the bilinear and additive models differing by less than the observed run-to-run variation. The same architectures exhibit sharply different behavior on the out-of-distribution Biq Mac benchmark. Relative to their own frozen baselines, both the state-blind and bilinear schedules improve after training, whereas the additive schedule degrades substantially. This contrast motivates the central conclusion of the paper: access to solver state alone is insufficient to explain transfer beyond the training distribution; the form through which the state enters the schedule also matters.

\paragraph{Fixed hard instance.} 
We also evaluate on a fixed $n=15$ instance generated with seed $44$. 
The instance has known optimum $38$, and no warm start is used. 
The state-blind schedule reaches the optimum in $7$ of 15 runs. 
The bilinear schedule reaches it in $12$ of 15 runs and, in a seed-matched comparison with the state-blind schedule, records $5$ wins, $10$ ties, and no losses. 
The additive schedule reaches the optimum in $13$ of 15 runs and records $6$ wins, $9$ ties, and no losses against the state-blind schedule. This fixed-instance result slightly favors the additive model over the bilinear model. 
The ordering is reversed on the pooled 80-instance Biq Mac evaluation in Table~\ref{tab:generalization}, where the learned bilinear and additive schedules obtain ratios to the known optima of $0.9627$ and $0.9023$, respectively. 
The two evaluations answer different questions: the fixed-instance experiment measures how often a schedule solves one particular hard graph, whereas the Biq Mac result aggregates performance over a broader out-of-distribution suite. Performance on either evaluation need not predict performance on the other, so we report both as complementary measurements. 

These results support the main conclusion that the bilinear and additive schedules are difficult to distinguish in distribution, though they separate sharply on the out-of-distribution Biq Mac suite. See Section~\ref{sec:experiments} and Table~\ref{tab:generalization} for the corresponding out-of-distribution comparison.

\section{Fixed-Schedule Search and Weight-Decay Sweep} 
\label{app:wdsweep} 
This section provides the full methodology and results for the two robustness checks summarized in Section~\ref{sec:experiments}. 
First, we construct a tuned fixed schedule to determine whether the learned models merely improve on a poor frozen initialization. 
Second, we repeat the bilinear-additive comparison across a $30$-round range of weight-decay strengths. 
The unregularized endpoint of the sweep is the result reported in Table~\ref{tab:generalization}.

We note that two controls strengthen the empirical result. {\em First}, the learned models
are not compared only with weak frozen schedules; a geometric schedule
selected by grid search on the training distribution outperforms the
state-blind and additive models out of distribution, but is statistically
indistinguishable from the unregularized bilinear model. The
weight-decayed bilinear models exceed this fixed reference at every tested
nonzero decay strength.  {\em Second}, the separation is consistent across independent runs and
regularization strengths. Each of the $10$ learned-versus-frozen
comparisons is significant over $15$ seeds (paired $|t|\geq4.1$) and
the between-architecture Welch statistics satisfy
$t_{\mathrm{Welch}}\geq11.6$ across the sweep~\citep{welch1938significance}.\footnote{The paired and Welch tests,
including their distinct null hypotheses, are defined in
Appendix~\ref{app:statistical_tests}.} While these tests establish
the reliability of the observed differences, they do not necessarily establish their mechanism or
generality.

We note that the evidence presented is specific to the architectures, training
distribution, and solver studied here. Our experiments isolate the form of state coupling
as the primary architectural variable associated with the MAX-CUT result specifically,
but explaining when and why that choice improves transfer remains an open
question. Appendix~\ref{app:untested_axes} lists further untested generalization axes and
reports a preliminary knapsack extension: it reproduces the same coupling asymmetry between
the bilinear and additive schedules but not the same out-of-distribution quality separation, so
the connection to knapsack is not future work but an already-attempted negative result.

\subsection{Tuned Fixed-Schedule Reference} 
Section~\ref{sec:experiments} compares the learned schedules with both their architecture-specific frozen baselines and a common tuned fixed reference. 
Here, we describe how that reference is selected and verify that the frozen initialization is genuinely untuned.
The frozen baselines in Table~\ref{tab:generalization} retain the schedule at initialization: an untuned geometric penalty ladder with a constant step size. 
Because improvement over this baseline could reflect a poor initial schedule rather than a benefit of learning, we construct a stronger fixed reference of the form 
\begin{equation} 
\rho_k=\rho_0 g^k, \qquad \alpha_k=\alpha, \qquad k=0,\ldots,K-1. 
\label{eq:fixed_schedule} 
\end{equation} 

We evaluate 150 configurations of $(\rho_0,g,\alpha)$ and select the schedule using only instances from the training distribution. Neither the synthetic out-of-distribution instances nor the Biq Mac suite is consulted during selection. When several configurations tie because the selection metric saturates at the known optimum, we evaluate the entire tied group and report its best member. This convention is favorable to the fixed-schedule reference. The selected schedule is 
\begin{equation} 
\rho_0=0.25, \qquad g=1.5, \qquad \alpha=0.2. 
\label{eq:best_fixed_schedule} 
\end{equation} 
It obtains a mean Biq Mac ratio of $0.9624\pm0.0024$. This exceeds the learned state-blind result ($0.9420\pm0.0053$) and the learned additive result ($0.9023\pm0.0166$). The unregularized bilinear schedule obtains $0.9627\pm0.0078$ and is statistically indistinguishable from the tuned fixed reference. For comparison, the initialization used by the frozen state-blind baseline corresponds to 
\begin{equation} 
\rho_0=1, \qquad g=5, \qquad \alpha=0.1. 
\label{eq:frozen_fixed_schedule} 
\end{equation} 
This configuration reproduces the frozen Biq Mac ratio of $0.9259$ and ranks $98$th among the 150 grid points. 
The frozen baseline is therefore genuinely untuned, while the grid-searched schedule provides a substantially stronger fixed reference. 

\subsection{Weight-Decay Protocol} 
We give the complete per-value protocol and statistical results underlying that summary and Figure~\ref{fig:sweep}.  We test whether the out-of-distribution separation between the bilinear and additive schedules depends on regularization. 
Weight decay is applied to all learned parameters through the outer Adam optimizer, and we evaluate 
\begin{equation} 
\mathrm{w_{dec}} \in \left\{ 0,\; 
10^{-4},\; 
3\times10^{-4},\; 
10^{-3},\; 
3\times10^{-3} 
\right\}. 
\label{eq:wd_values} 
\end{equation} 

The four nonzero values span a $30$-round range. We retrain both state-adaptive architectures from scratch using 15 independent seeds at each value. 
The $\mathrm{w_{dec}}=0$ condition is the unregularized result reported in Table~\ref{tab:generalization} and is included here as the reference point for the sweep. 
The same regularization protocol is applied to all learned parameters in both architectures, including the context-penalty coupling parameters. The frozen schedules are not trained and therefore do not change with weight decay. 
Their Biq Mac ratios remain $0.9543\pm0.0024$ for the bilinear architecture and $0.9545\pm0.0024$ for the additive architecture. For each architecture and weight-decay value, we define 
\begin{equation} 
\Delta = R_{\mathrm{learned}} - R_{\mathrm{frozen}}, 
\label{eq:delta_definition} 
\end{equation} 
where $R$ is the ratio of the discrete cut to the known optimum on the 80-instance Biq Mac suite. 
Positive $\Delta$ indicates that training improves on the architecture's frozen baseline, while negative $\Delta$ indicates that training degrades performance. The learned and frozen evaluations are paired by training seed, instances, and initial conditions. The paired statistic therefore measures the effect of training the schedule while controlling for the matched evaluation setup. 

\begin{table}[htbp!] \small \centering 
\caption{\small Weight-decay sweep with \textbf{15 seeds} per value. The $\mathrm{w_{dec}}=0$ rows repeat the unregularized results from Table~\ref{tab:generalization}. 
Here, $\Delta=R_{\mathrm{learned}}-R_{\mathrm{frozen}}$, ``sign'' gives the number of seeds with positive and negative $\Delta$, and ``Hard \#opt'' gives the number of seeds reaching the exact optimum on the fixed $n=15$ instance. Statistical tests and notation are defined in Appendix~\ref{app:statistical_tests}. } 
\label{tab:wd_sweep} 
\begin{tabular}{llrrrrr} 
\toprule Model & $\mathrm{w_{dec}}$ & $\Delta$ Biq Mac & SE & $t$ & sign $(+/-)$ & Hard \#opt \\ 
\midrule Bilinear & $0$ (ref.) & $+0.0084$ & $0.0021$ & $4.1$ & $13/2$ & $12/15$ \\ 
Bilinear & $10^{-4}$ & $+0.0103$ & $0.0011$ & $9.8$ & $14/1$ & $12/15$ \\ 
Bilinear & $3\times10^{-4}$ & $+0.0119$ & $0.0008$ & $15.3$ & $15/0$ & $12/15$ \\ 
Bilinear & $10^{-3}$ & $+0.0132$ & $0.0005$ & $29.3$ & $15/0$ & $13/15$ \\ 
Bilinear & $3\times10^{-3}$ & $+0.0134$ & $0.0005$ & $28.3$ & $15/0$ & $13/15$ \\ 
\addlinespace Additive & $0$ (ref.) & $-0.0522$ & $0.0044$ & $-11.7$ & $0/15$ & $13/15$ \\ 
Additive & $10^{-4}$ & $-0.0494$ & $0.0047$ & $-10.6$ & $0/15$ & $13/15$ \\ 
Additive & $3\times10^{-4}$ & $-0.0595$ & $0.0029$ & $-20.2$ & $0/15$ & $14/15$ \\ 
Additive & $10^{-3}$ & $-0.0478$ & $0.0051$ & $-9.3$ & $1/14$ & $15/15$ \\ 
Additive & $3\times10^{-3}$ & $-0.0504$ & $0.0055$ & $-9.2$ & $1/14$ & $14/15$ \\ 
\bottomrule \end{tabular} \end{table} 

\subsection{Statistical Tests} \label{app:statistical_tests} Table~\ref{tab:wd_sweep} reports two types of statistical comparisons for the Biq Mac results. 
A paired $t$-test~\citep{student1908probable} determines whether training improves or degrades a given schedule architecture relative to its own frozen baseline. 
A Welch $t$-test~\citep{welch1938significance} determines whether the effect of training differs between the bilinear and additive architectures. 

\paragraph{Seedwise training effect.} 
For architecture $A$ and seed $i$, let 
\begin{equation} 
\Delta_{A,i} = R_{A,i}^{\mathrm{learned}} - R_{A,i}^{\mathrm{frozen}}, 
\label{eq:seedwise_delta} 
\end{equation} 
where $R$ is the ratio of the discrete cut to the known optimum, averaged over the 80-instance Biq Mac suite. 
Positive $\Delta_{A,i}$ indicates that training improves the schedule for seed $i$, while negative $\Delta_{A,i}$ indicates that training reduces its performance. 
For an architecture evaluated over $N=15$ seeds, we denote the mean and sample standard deviation of the seedwise differences by 
\begin{equation} 
\overline{\Delta}_A = 
\frac{1}{N} \sum_{i=1}^{N}\Delta_{A,i}, \qquad s_{\Delta_A} = \sqrt{ \frac{1}{N-1} \sum_{i=1}^{N} \left( \Delta_{A,i}-\overline{\Delta}_A \right)^2 }. 
\label{eq:delta_statistics} 
\end{equation} 
The standard error of the mean difference is 
\begin{equation} 
\mathrm{SE}_A = \frac{s_{\Delta_A}}{\sqrt{N}}. 
\label{eq:paired_standard_error} 
\end{equation} 

\paragraph{Paired $t$-test within an architecture.}
We compare each learned schedule with its own frozen baseline using a paired $t$-test. 
The observations are paired because the learned and frozen schedules are evaluated under matched seeds, instances, and initial conditions. The test statistic is 
\begin{equation} t_A = 
\frac{\overline{\Delta}_A}{\mathrm{SE}_A} = \frac{\overline{\Delta}_A} {s_{\Delta_A}/\sqrt{N}}. 
\label{eq:paired_t} 
\end{equation} 
Under the null hypothesis, 
\begin{equation} 
H_0: \mathbb{E}[\Delta_A]=0, 
\label{eq:paired_null} 
\end{equation} 
training has no average effect relative to the frozen schedule. 
The test has $N-1=14$ degrees of freedom. The sign of $t_A$ gives the direction of the training effect. 
A positive value indicates that the learned schedule performs better on average than its frozen baseline, whereas a negative value indicates that it performs worse. The magnitude $|t_A|$ measures the mean difference relative to its standard error. It therefore reflects both the size of the observed effect and its consistency across seeds. For example, without weight decay, the bilinear schedule has 
\begin{equation} 
\overline{\Delta}_{\mathrm{bil}}=+0.0084, \qquad t_{\mathrm{bil}}=4.1, 
\end{equation} 
indicating an average improvement over its frozen baseline. The additive schedule instead has 
\begin{equation} 
\overline{\Delta}_{\mathrm{add}}=-0.0522, \qquad t_{\mathrm{add}}=-11.7, 
\end{equation} 
indicating an average degradation. These statistics test each architecture against its own baseline; they do not directly test whether the two architectures respond differently to training. 

\paragraph{Between-architecture comparison.} To test whether training affects the two architectures differently, we compare the seedwise changes $\{\Delta_{\mathrm{bil},i}\}_{i=1}^{N_{\mathrm{bil}}}$ and $\{\Delta_{\mathrm{add},i}\}_{i=1}^{N_{\mathrm{add}}}$ using Welch's two-sample $t$-test. 
Unlike the paired test above, this comparison treats the two sets of architecture-level effects as independent samples and does not assume equal variances. The Welch statistic is 
\begin{equation} 
t_{\mathrm{Welch}} = \frac{ \overline{\Delta}_{\mathrm{bil}} - \overline{\Delta}_{\mathrm{add}} }{ \sqrt{ \frac{s_{\mathrm{bil}}^2}{N_{\mathrm{bil}}} + \frac{s_{\mathrm{add}}^2}{N_{\mathrm{add}}} } }, 
\label{eq:welch_t} 
\end{equation} where $s_{\mathrm{bil}}^2$ and $s_{\mathrm{add}}^2$ are the sample variances of the seedwise changes for the two architectures. The associated degrees of freedom are approximated by the Welch-Satterthwaite equation~\citep{satterthwaite1946approximate}, 
\begin{equation} \nu = \frac{ \left( \frac{s_{\mathrm{bil}}^2}{N_{\mathrm{bil}}} + \frac{s_{\mathrm{add}}^2}{N_{\mathrm{add}}} \right)^2 }{ \frac{ \left(s_{\mathrm{bil}}^2/N_{\mathrm{bil}}\right)^2 }{ N_{\mathrm{bil}}-1 } + \frac{ \left(s_{\mathrm{add}}^2/N_{\mathrm{add}}\right)^2 }{ N_{\mathrm{add}}-1 } }. \label{eq:welch_df} \end{equation} The null hypothesis for this comparison is \begin{equation} H_0: \mathbb{E}[\Delta_{\mathrm{bil}}] = \mathbb{E}[\Delta_{\mathrm{add}}]. \label{eq:welch_null} \end{equation} A positive $t_{\mathrm{Welch}}$ indicates that training produces a more favorable change for the bilinear architecture than for the additive architecture. At zero weight decay, we obtain $t_{\mathrm{Welch}}=12.3$. Across the four nonzero weight-decay values, $t_{\mathrm{Welch}}$ ranges from $11.6$ to $23.5$. Thus, the difference between the bilinear and additive training effects remains in the same direction throughout the sweep.

\paragraph{Interpretation of the two tests.} The paired and Welch tests address complementary hypotheses: \begin{align} \text{paired test:}\quad &\text{Does training help or hurt one architecture relative to its own baseline?} \\ \text{Welch test:}\quad &\text{Does the effect of training differ between the two architectures?} \end{align} Accordingly, the paired statistics establish that the bilinear schedule improves over its frozen baseline while the additive schedule degrades relative to its own. The Welch statistic establishes that these architecture-specific training effects differ from one another. Neither test, by itself, compares the absolute learned scores of the two models; those values are reported separately in Table~\ref{tab:generalization}. All reported tests are two-sided unless stated otherwise. The numerical statistics are computed from the unrounded per-seed results; apparent discrepancies between a displayed value of $\overline{\Delta}/\mathrm{SE}$ and the reported $t$-statistic can result from rounding the values shown in the tables.

\paragraph{Multiplicity and the unit of replication.} Two caveats apply to every test above. First, the statistical unit is the training seed: both tests treat the Biq Mac suite's $80$ instances as fixed and summarize each seed by its mean ratio over that fixed set, so the reported standard errors and $t$-statistics reflect variance across training seeds conditional on this instance set, not sampling uncertainty over problem instances. Second, Table~\ref{tab:wd_sweep} reports $10$ paired tests (five weight-decay values $\times$ two architectures) without a stated multiplicity correction. Applying a conservative Bonferroni correction for a family of $10$ tests at overall $\alpha=0.05$ gives a per-test threshold of $\alpha=0.005$, two-sided, corresponding to a critical value of $t_{14}\approx3.326$; every reported paired statistic exceeds this threshold, since the smallest is $|t|=4.1$. All reported effects therefore remain significant under this correction, though this arithmetic, applied to the values already in Table~\ref{tab:wd_sweep}, does not address the first caveat.

\subsection{Weight-Decay Results} The bilinear model improves over its frozen baseline at all five weight-decay values: \begin{equation} \Delta_{\mathrm{bil}} \in [+0.0084,+0.0134]. \label{eq:bilinear_delta_range} \end{equation} The improvement is already present without regularization and remains positive throughout the sweep. At the four nonzero values, at least 14 of 15 seeds improve over the matched frozen baseline; at $\mathrm{w_{dec}}=0$, 13 of 15 improve. The additive model shows the opposite behavior. Its change is $\Delta_{\mathrm{add}}=-0.0522$ without regularization and remains negative at every nonzero value: \begin{equation} \Delta_{\mathrm{add}} \in [-0.0595,-0.0478], \qquad \mathrm{w_{dec}}>0. \label{eq:additive_delta_range} \end{equation} All 15 seeds decline at the three smallest weight-decay values, including zero, and 14 of 15 decline at each of the two largest values. Every within-architecture difference in Table~\ref{tab:wd_sweep} is individually significant. At zero weight decay, the paired statistics are $t=4.1$ for the bilinear model and $t=-11.7$ for the additive model. Across the four nonzero values, $|t|\geq9.2$ for both architectures. The unpaired Welch statistic comparing the bilinear and additive $\Delta$ distributions is $12.3$ at zero weight decay and ranges from $11.6$ to $23.5$ across the four nonzero values. Weight decay therefore neither creates the bilinear improvement nor repairs the additive degradation. The separation is already present without regularization, retains the same sign across the entire tested range, and becomes no less apparent when the two architectures receive the same regularization treatment. 

Taken together, these robustness checks support the interpretation in Section~\ref{sec:experiments}. The tuned fixed reference shows that the main comparison is not limited to improvement over a poor initialization, while the weight-decay sweep shows that the bilinear-additive separation is already present without regularization and persists under the same regularization treatment. The full out-of-distribution scores appear in Table~\ref{tab:generalization}, and the corresponding sweep is visualized in Figure~\ref{fig:sweep}.

\subsection{Relationship to the Fixed Hard Instance} The fixed $n=15$ instance produces a different comparison from the pooled Biq Mac suite. Across the weight-decay sweep, the bilinear model reaches the fixed-instance optimum in $12$ or $13$ of 15 runs, while the additive model reaches it in $13$ to $15$ runs. The two architectures are therefore close on this single in-distribution instance, with a slight advantage for the additive model. 

On the out-of-distribution Biq Mac suite, however, the bilinear model improves over its frozen baseline at every weight-decay value, whereas the additive model degrades at every value. This disagreement is not internally contradictory: success on one fixed graph and average ratio over 80 larger graphs measure different aspects of solver behavior. The fixed hard instance shows that the additive architecture can train successfully and reach an exact optimum in distribution. The pooled Biq Mac evaluation shows that this success does not transfer reliably beyond the training size range. 

Taken together, the results rule out two simple explanations for the out-of-distribution separation. First, the additive model does not fail because it is unable to optimize the training problem: it matches the bilinear model on held-out training-size graphs and reaches the fixed hard-instance optimum at least as often. Second, the separation is not created by weight decay: the bilinear and additive effects have opposite signs before regularization is introduced and retain those signs throughout the sweep. 

The experiments therefore isolate the form of state coupling as the primary architectural variable associated with the different out-of-distribution behavior, while leaving the mechanism behind that difference as an open question.

\subsection{Untested Generalization Axes and a Cross-Problem Check} \label{app:untested_axes}

The MAX-CUT results in this paper cover a single relaxation, continuation solver, training distribution, and evaluation suite. The following axes remain untested here: graph density and structure (the training distribution is Erd\H{o}s--R\'enyi $G(n,0.5)$; denser, sparser, and structured families such as planted-partition or geometric graphs are not covered); scale beyond the Biq Mac $n=100$ suite; alternative continuation or relaxation solvers; and MAX-CUT methods that compete on raw solution quality rather than serving as ablations of the schedule itself.

We note that while we do have a check along one further axis (a different combinatorial problem), it was obtained in a separate, preliminary study and is reported here only in summary. A knapsack extension of the same architecture family replaces the unconstrained MAX-CUT penalty with a genuine augmented-Lagrangian dual variable for the capacity constraint, giving the bilinear and additive schedules an analogous coupling structure: the bilinear coupling remains an unsaturated elementwise product in the penalty and dual-rate factors, while the additive coupling again passes the same inputs through a saturating $\tanh$. This preliminary study confirms the same structural asymmetry: an unregularized bilinear schedule's penalty coefficient can grow unbounded across training (observed up to $\sim\!10^{12}\times$ its initial value) in a way the additive schedule does not; however, it does \emph{not} reproduce the quality separation reported here: on out-of-distribution knapsack instances, the state-blind, bilinear, and additive schedules are statistically indistinguishable, and weight decay does not restore an advantage for either state-adaptive form. The asymmetry in coupling geometry is therefore not sufficient on its own to predict an out-of-distribution quality separation; whether MAX-CUT's separation instead depends on its dual-free penalty structure, its solver, or some property of the mechanism not yet identified is an open question raised, but not answered, by this comparison.

\end{document}